\documentclass[sigplan,screen]{acmart}

\AtBeginDocument{%
  }

\acmDOI{10.1145/nnn.nnn} 
\acmISBN{978-x-xxxx-xxxx-x/YY/MM} 
\acmConference[GECCO '24]{The Genetic and Evolutionary Computation Conference 2024}{July 14--18, 2024}{Melbourne, Australia}
\acmYear{2024}
\copyrightyear{2024}
\acmISBN{978-1-4503-XXXX-X/18/06}

\begin{document}
\title{Exploring the Improvement of Evolutionary Computation via Large Language Models}

\author{Jinyu Cai}
\affiliation{%
  \institution{Waseda University}
  \city{Tokyo}
  \country{Japan}}
\email{bluelink@toki.waseda.jp}

\author{Jinglue Xu}
\affiliation{%
  \institution{University of Tokyo}
  \city{Tokyo}
  \country{Japan}}
\email{jingluexu@gmail.com}

\author{Jialong Li}
\authornote{Corresponding Author: Jialong Li}
\affiliation{
  \institution{Waseda University}
  \city{Tokyo}
  \country{Japan}}
\email{lijialong@fuji.waseda.jp}

\author{Takuto Yamauchi}
\affiliation{
  \institution{Waseda University}
  \city{Tokyo}
  \country{Japan}}
\email{itaku3@akane.waseda.jp}

\author{Hitoshi Iba}
\affiliation{
  \institution{University of Tokyo}
  \city{Tokyo}
  \country{Japan}}
\email{iba@iba.t.u-tokyo.ac.jp}

\author{Kenji Tei}
\affiliation{
  \institution{Tokyo Institute of Technology}
  \city{Tokyo}
  \country{Japan}}
\email{tei@c.titech.ac.jp}

\renewcommand{\shortauthors}{J Li. et al.}

\begin{abstract}
Evolutionary computation (EC), as a powerful optimization algorithm, has been applied across various domains. However, as the complexity of problems increases, the limitations of EC have become more apparent. The advent of large language models (LLMs) has not only transformed natural language processing but also extended their capabilities to diverse fields.
By harnessing LLMs' vast knowledge and adaptive capabilities, we provide a forward-looking overview of potential improvements LLMs can bring to EC, focusing on the algorithms themselves, population design, and additional enhancements. This presents a promising direction for future research at the intersection of LLMs and EC.
\end{abstract}

\maketitle

\section{Introduction}

Evolutionary Computation (EC)—derived from principles of natural selection and genetic algorithms—is employed extensively in robotics, bioinformatics, and AI prediction. EC methodologies simulate the iterative evolution of populations through selection, crossover, and mutation, enabling efficient explorations of complex solution spaces. 
However, EC still faces significant challenges~\cite{zhan2022survey}, such as (i) handling large search spaces and complex objective functions, which can trap algorithms in local optima and escalate experimental and computational costs. (ii) Evolutionary algorithms (EAs) often require meticulous human design and domain expertise to perform optimally. The configuration of EAs is dependent on a thorough understanding of the problem structure and the creation of specific evolutionary operators, significantly limiting their applicability.

In recent years, large language models (LLMs), represented by ChatGPT, have sparked a revolution in language processing technology. These models, with their prior knowledge and exceptional language understanding abilities, have demonstrated significant potential in various domains such as writing, programming, and optimization.
With these capabilities, LLMs hold the potential to ameliorate some of the prevalent shortcomings in current EC strategies. Specifically, LLMs can leverage their extensive knowledge and context understanding to reduce the search space and guide evolutionary processes~\cite{tornede2023automl}. Additionally, by integrating LLMs capabilities into evolutionary operators~\cite{lanzi2023chatgpt,chen2024evoprompting}, we can substantially enhance the versatility and scope of EC, particularly when tackling complex and ambiguously defined language-related tasks.


This paper aims to explore how to address these challenges by integrating LLMs in the era of LLMs, starting from the main challenges and limitations currently faced by EC, and to look forward to future development directions. We divide the main directions of LLM-driven EC into three parts:
(i)LLM-driven EA algorithms: Focusing on EA algorithms, we explore how LLMs can improve the core mechanisms of EA, including selecting evolutionary strategies and improving evolutionary operators.
(ii)LLM-driven population and individuals: Regarding individuals and populations, we investigate how LLMs can optimize the design of populations and the generation and evaluation processes of individuals.
(iii)Other improvements: We examine the role of LLMs in expanding the application areas of EC, enhancing user interaction, and adapting to dynamic environments.

\section{LLM-driven EA Algorithms}
With the continuous development of information technology, the optimization problems that EAs need to solve are becoming increasingly complex. These complexities often manifest in the large scale of the problems, their multi-objectivity, and the existence of numerous local/global optima. Traditional evolutionary algorithms may encounter difficulties when addressing these complex challenges. By leveraging LLMs, researchers could potentially guide evolutionary algorithms more effectively to address these challenges.


\textbf{LLM-assisted Evolutionary Strategy Selection}: For complex optimization challenges, it's vital to adopt specific strategies for different issues. Traditionally, strategy determination relies on researchers' experience, which can be inaccurate and inefficient given the problem's complexity and vast search space. LLMs provide a novel approach by analyzing problems and interpreting data, offering strategy recommendations based on historical performance and problem characteristics to guide EA.
Among them,~\cite{liu2023large} introduced an innovative method called LLM-driven EA (LMEA) for solving combinatorial optimization problems. In each generation of evolutionary search, LMEA guides LLMs to select parent solutions from the population and carry out crossover and mutation operations to generate offspring solutions. Then, LLMs assess these new solutions and incorporate them into the next generation of the population. Experimental results show that LMEA demonstrates superior performance in handling instances of the Traveling Salesman Problem compared to traditional heuristic methods, with significantly lower demands on researchers' expertise.

\textbf{Evolutionary Operators via LLMs}: In EC, the design of evolutionary operators plays a crucial role. Traditional evolutionary operators are usually fixed and come with explicit mathematical definitions. Introducing LLMs can break this fixed mindset, expanding EC to a broader range. We categorize the impact of LLMs on evolutionary operators into the following two types:

\textit{LLM as an Evolutionary Operator}: In this scenario, LLMs directly act as an evolutionary operator involved in the generation, modification, or selection of individuals. This change breaks the fixed mathematical definition of traditional evolutionary operators, allowing evolutionary algorithms to be better applied in different fields, such as text~\cite{lanzi2023chatgpt} and code generation~\cite{chen2024evoprompting}. Specifically, these studies are characterized by, but not limited to, leveraging LLMs for operations including generation, crossover, and mutation.

\textit{LLM-guided Evolutionary Operators}: Here, LLMs are used to guide genetic operators, such as crossover and mutation, based on the task scenario and objectives, making the operations more effective. For example,~\cite{ye2024reevo} uses LLMs to guide the generation and optimization of algorithms, including dynamic adjustment of hyperparameters.

\textit{LLM-generated Evolutionary Operators}: As a prospect, we believe that based on the above, LLMs can also directly generate new or modified operators, providing unprecedented genetic operation methods and exploring new areas of the solution space.

\section{LLM Improvement of Populations and Individuals}
In modern EC, particularly with problems involving high levels of abstraction and text-based complexity, the traditional EC methods often fall short. These issues typically contain rich and complex data, challenging the optimal performance of conventional approaches. The integration of LLMs offers promising avenues for enhancing the efficiency and applicability of EC in complex domains.

\textbf{LLM Assistance in Handling Complex High-Dimensional Data}: LLMs, with their exceptional logical thinking abilities, can process and understand high-dimensional and highly abstract data~\cite{maddigan2024explaining}. By leveraging LLMs to extract key features and patterns from extensive datasets, we can substantially optimize the input for evolutionary algorithms, thereby enhancing the efficiency of data processing and the quality of final solutions.

\textbf{LLM Strengthening of Population Design}: Utilizing LLMs to design or refine population structures and initialization ensures diversity and high-quality starting points, thus accelerating the convergence speed and improving the quality of solutions. For instance, studies like~\cite{liu2023algorithm} have applied EC to algorithmic optimization tasks, integrating LLMs' powerful programming capabilities to generate related optimization algorithms as populations within the EC framework, achieving results that surpass traditional methods.

\section{Other Improvements and Limitations}

Beyond directly strengthening the core mechanisms and characteristics of EC, LLMs can also enhance EC's functionality in broader domains. 

\textbf{Multimodal LLM Expansion of EC Applications}: By integrating multimodal LLMs, such as combining text, image, and sound information, the application areas of EC can be significantly expanded. For example, using textual information to guide algorithms in finding solutions or employing images to evaluate individual fitness can greatly enhance the applicability and effectiveness of EC methods across different fields.

\textbf{Interactive EC and Adaptation to Dynamic Environments}: Integrating natural language processing from LLMs with EC algorithms improves human-algorithm interaction, utilizing human input for optimization. LLMs' extensive knowledge enhances adaptability to various changes and challenges. For instance, as shown in LMEA~\cite{liu2023large}, EC can swiftly adapt to diverse optimization tasks by updating prompt descriptions and instructions, highlighting LLMs' adaptability in dynamic settings.

Despite these advancements, LLMs also present intrinsic limitations. As noted in~\cite{yang2023large}, LLMs as optimizers are suited primarily for less complex issues and may underperform relative to basic optimization algorithms when faced with increased complexity. 
Furthermore, as~\cite{tornede2023automl} discusses, LLMs performance heavily relies on prompt design and contextual constraints. Inadequacies in prompt formulation or contextual framing can result in performance fluctuations, underscoring the dependency of LLMs on input specifications.

\section{Conclusion}
This paper explores how to strengthen traditional EC methods by integrating them with LLMs in the era of LLMs. We explored the achievements and potential of LLMs in improving evolutionary strategy selection, optimizing population design, and enhancing evolutionary operators. We point out that combining EC with LLMs can help address traditional challenges faced by EC, such as the ability to handle high-dimensional and complex problems and breakthroughs in algorithm automation and interactive EC. We hope this paper can bring new inspiration and insights to the EC community.


\bibliographystyle{ACM-Reference-Format}
\bibliography{main}

\end{document}